\documentclass[11pt]{article}

\usepackage{cite}
\usepackage{amsmath,amssymb,amsfonts}
\usepackage{bm,mathtools}
\usepackage{algorithmic}
\usepackage{graphicx}
\usepackage{algorithm}
\usepackage{hyperref}
\hypersetup{
    colorlinks=true,
    linkcolor=blue,
    filecolor=magenta,      
    urlcolor=cyan,
    citecolor=blue
}
\usepackage{textcomp}
\usepackage{booktabs}
\usepackage{multirow}
\usepackage{float}
\usepackage{enumitem}

\usepackage[margin=1in]{geometry}

\DeclareTextFontCommand{\textbf}{\bfseries}
\DeclareTextFontCommand{\textit}{\itshape}

\newcommand{\R}{\mathbb{R}}

\newcommand{\Kset}{\mathcal{K}}
\newcommand{\Sset}{\mathcal{S}}

\begin{document}

\title{Joint Score-Threshold Optimization for Interpretable Risk Assessment}

\author{
Fardin Ganjkhanloo\thanks{F. Ganjkhanloo and E. Springer contributed equally to this work. This work was supported by the Doctors Company Foundation. Corresponding author: K. Ghobadi.}\,$^{1,2}$, 
Emmett Springer\footnotemark[1]\,$^{3,2}$, 
Erik H. Hoyer$^{4,5}$, 
Daniel L. Young$^{4,6}$, 
Kimia Ghobadi$^{3,2}$ \\
\\
\small $^1$Center for Health Systems and Policy Modeling, Department of Health Policy and Management, \\
\small Johns Hopkins University \\
\small $^2$Malone Center for Engineering in Healthcare, Johns Hopkins University \\
\small $^3$Center for Systems Science and Engineering, Department of Civil and Systems Engineering, \\
\small Johns Hopkins University \\
\small $^4$Department of Physical Medicine and Rehabilitation, School of Medicine, Johns Hopkins University \\
\small $^5$Johns Hopkins Hospital \\
\small $^6$Department of Physical Therapy, University of Nevada, Las Vegas \\
\\
\small \texttt{fganjkh1@jhu.edu, espring6@jh.edu, kimia@jhu.edu, ehoyer@jhmi.edu, daniel.young@unlv.edu}
}

\date{}

\maketitle

\begin{abstract}
Risk assessment tools in healthcare commonly employ point-based scoring systems that map patients to ordinal risk categories via thresholds. While electronic health record (EHR) data presents opportunities for data-driven optimization of these tools, two fundamental challenges impede standard supervised learning: (1) labels are often available only for extreme risk categories due to intervention-censored outcomes, and (2) misclassification cost is asymmetric and increases with ordinal distance. We propose a mixed-integer programming (MIP) framework that jointly optimizes scoring weights and category thresholds in the face of these challenges. Our approach prevents label-scarce category collapse via threshold constraints, and utilizes an asymmetric, distance-aware objective. The MIP framework supports governance constraints, including sign restrictions, sparsity, and minimal modifications to incumbent tools, ensuring practical deployability in clinical workflows. We further develop a continuous relaxation of the MIP problem to provide warm-start solutions for more efficient MIP optimization. We apply the proposed score optimization framework to a case study of inpatient falls risk assessment using the Johns Hopkins Fall Risk Assessment Tool.
\end{abstract}

\noindent\textbf{Keywords:} Clinical decision support, fall risk assessment, interpretable machine learning, mixed-integer programming, ordinal classification, risk scoring systems.

\section{Introduction}
\label{sec:intro}

Across healthcare settings, risk assessment is routinely implemented via itemized, checklist-style instruments in which clinicians mark the presence, absence, or severity of predefined risk factors. Each factor contributes a predetermined integer point value to a total score, and pre-specified integer thresholds map patient scores to ordinal risk categories (e.g., Low, Medium, High) that govern downstream clinical workflows such as monitoring frequency or preventive interventions \cite{poe2007johns, oliver1997development, jette2014pac, caprini1991clinical, bergstrom1987braden}. The restriction to integer coefficients and thresholds is not merely conventional but essential for practical deployment, enabling rapid bedside assessment and facilitating clinician trust through cognitive transparency. This point-based scoring paradigm is widely adopted due to its transparency and ease of use: clinicians can compute risk scores through simple mental calculations without requiring calculators or computational tools \cite{rudin2019stop, ustun2019learning, holzinger2017we}. However, many widely used point-based risk scoring tools were developed based on clinical reasoning rather than data-driven evidence, and warrant updating for current clinical practice and changing populations \cite{vincent_sofa_1996, moreno_sequential_2023}.

The increasing availability of granular electronic health record (EHR) data presents an opportunity to optimize such tools through data-driven methods: re-estimating integer point weights, adjusting integer thresholds, and encoding deployment constraints while preserving both the interpretable structure and computational simplicity that clinicians trust \cite{goldstein2016opportunities, rajkomar2019machine}. The practical realities of clinical deployment introduce two fundamental challenges that violate the assumptions of standard supervised learning approaches. The first challenge is a frequent lack of ground-truth labels for patients' level of risk. In deployment, patient outcomes and clinical interventions are inherently coupled, preventing clear causal analysis between risk factors and outcomes. The second challenge is with promoting model assuredness and alignment with clinical priorities. Unconstrained supervised learning can result in unexpected outcomes that do not align with the realities of available resources or clinical objectives. Operationally, under-triage (classifying a truly high-risk patient as low-risk) can result in preventable adverse events with serious consequences, while over-triage (classifying a low-risk patient as high-risk) can result in resource inefficiency, alert fatigue, and worse patient outcomes. The relative severity of under- versus over-triage depends on the specific adverse event, available interventions, and patient population. While some existing models can balance this tradeoff via model hyperparameters, they are not equipped to safely control the resulting risk score distributions.

We introduce a constrained mixed-integer optimization framework for patient risk classification that addresses these challenges while preserving the interpretable point-score structure essential for clinical adoption. Our approach handles settings where labels exist only for extreme risk categories. In the extreme case (which applies in our clinical application), intervention-censored outcomes prevent reliable labeling of middle-risk categories entirely. We can confidently identify only the extremes: patients who experienced adverse events despite interventions (high-risk) and those who remained event-free without resource-intensive interventions (low-risk). Patients who received interventions but remained event-free have maximally uncertain labels, as it is not possible to fully determine whether interventions prevented adverse events or were unnecessarily applied. For these ambiguous cases, we do not force a label and exclude them from training. We retain these cases for validation, comparing the distributions of scores between risk categories across models.

We embed misclassification costs that increase with ordinal distance and allow directional asymmetry, directly aligning the training objective with deployment priorities. When middle categories lack labeled examples, we prevent threshold collapse through minimum gap constraints derived from incumbent tools or cross-validation, maintaining clinically meaningful risk stratification. Finally, we extend ideas from smooth hinge losses \cite{rennie2005smooth} to the ordinal setting with our derived constrained score optimization (CSO) relaxation, which uses softplus losses to preserve the asymmetric, distance-aware training signal while enabling efficient optimization for large-scale problems. We develop a joint model optimization pipeline with MIP and CSO, and apply this methodology to a case study of improving the Johns Hopkins Fall Risk Assessment Tool (JHFRAT) for assessing inpatient fall risk.

The remainder of this paper is organized as follows. Section \ref{sec:related-work} highlights related work in the area of ordinal risk assessment with regression methods and constrained optimization. Section \ref{sec:methods} presents our mixed-integer programming formulation and convex relaxation. Section \ref{sec:experiments} describes experimental evaluation on clinical risk assessment tasks. Section \ref{sec:discussion} discusses computational considerations, limitations, and deployment aspects. Section \ref{sec:conclusion} concludes with implications for precision medicine.

\section{Related Work}
\label{sec:related-work}

Ordinal regression methods have been extensively studied, but existing approaches fail to address the unique challenges of clinical risk assessment. Classical ordinal regression methods such as the proportional odds model \cite{mccullagh1980regression} assume parallel decision boundaries and complete supervision across all categories. Support vector approaches for ordinal regression \cite{chu2007support} and threshold models \cite{shashua2002ranking} similarly require fully labeled training data. Recent extensions to deep learning for ordinal problems \cite{niu2016ordinal, cao2020rank} and specialized ranking losses \cite{rennie2005loss} maintain these restrictive assumptions, treating all categories as equally important and all misclassification costs as symmetric. The partial label learning literature addresses scenarios where each instance is associated with a set of candidate labels \cite{cour2011learning, wang2019partial}, with disambiguation methods that iteratively refine label assignments \cite{zhang2015solving, wang2019adaptive} and loss-based methods that modify training objectives \cite{feng2020provably, lv2020progressive}. However, our setting differs fundamentally in three ways: (1) feasible sets arise from systematic censoring due to clinical interventions rather than random labeling ambiguity, (2) misclassification costs are inherently asymmetric and distance-aware in healthcare applications, and (3) we require preservation of ordinal structure throughout the learning process \cite{karapanagiotis2023tailored}. Furthermore, while extensions like partial proportional odds models \cite{peterson1990partial} relax the proportional odds assumption, they still assume complete supervision and do not address the selective labeling problem where entire categories may lack reliable labels.

We build upon the literature for constrained mixed integer optimization methods as an alternative to ordinal regression. Early work employing integer programming methods for classification includes that of Uney and Turkay \cite{uney_mixed-integer_2006} and Bertsimas and Shioda \cite{bertsimas_classification_2007}. Both works solve supervised multi-class classification problems by separating different classes into polyhedral regions of the data space. Bertsimas and Shioda also solve a separate linear regression problem, which employs unsupervised classification to create different linear models across heterogeneous subpopulations. Billiet et al. \cite{billiet_interval_2018} formulate a risk-score weighting binary classification linear programming model based on soft-margin support vector machines. The Interval Coded Scoring linear program presented in that work utilizes similar model constraints as integer programming classification, but avoids the use of integer or binary variables via the SVM-based objective. Dedieu et al. \cite{dedieu_learning_2021} address the computational complexity of scaling integer programming classification models by solving a series of smaller relaxed MIP subproblems. Ustun and Rudin \cite{ustun2019learning} present the RiskSlim model, which optimizes integer risk model coefficients. They present both a nonlinear objective model and a surrogate linear program model. Among the current literature, their work is the most similar to our methodology, though their model applies only to binary risk classification. Furthermore, we use a discrete objective function to enforce class-based ordinal misclassification.

\section{Proposed Models}
\label{sec:methods}

We consider the problem of learning a risk stratification tool that maps patient features $x \in \mathbb{R}^p$ to ordinal risk categories $\mathcal{K} = \{1, 2, \ldots, K\}$ (e.g., Low, Medium, High for $K=3$). Following the clinical scoring system paradigm \cite{ustun2019learning}, the tool consists of integer-valued scoring weights $\beta \in \mathbb{Z}^p$ and integer thresholds $\tau = (\tau_1, \ldots, \tau_{K-1}) \in \mathbb{Z}^{K-1}$ with $\tau_1 < \tau_2 < \cdots < \tau_{K-1}$. The integer restriction enables clinicians to compute risk scores $s(x) = \beta^\top x$ through mental arithmetic without computational aids, a critical requirement for bedside deployment and cognitive transparency. A patient with score $s(x)$ is assigned to category $k$ if $\tau_{k-1} < s(x) \leq \tau_k$, where we define $\tau_0 = -\infty$ and $\tau_K = +\infty$ for notational convenience.

Given training data $\{(x_i, k_i^*, w_i)\}_{i \in \mathcal{I}}$ where $x_i \in \R^p$ are patient features, $k_i^* \in \Kset$ are known category labels, and $w_i > 0$ are instance weights, our goal is to learn integer parameters $(\beta, \tau) \in \mathbb{Z}^p \times \mathbb{Z}^{K-1}$ that minimize expected asymmetric ordinal loss subject to interpretability and operational constraints. Our setting considers the possibility that some patient samples will not have associated risk labels. Without imposition of assumptions or causal analysis of the effects of interventions, these samples must be excluded from training to maintain the integrity of the training labels. The key challenge then becomes \emph{partial supervision at the category level}: while each included training instance $i \in \mathcal{I}$ has a single known label $k_i^*$, certain categories $k \in \mathcal{K}$ may have no labeled training examples. Let $\mathcal{I}_k = \{i \in \mathcal{I} : k_i^* = k\}$ denote the set of training instances with known label $k$. In the extreme case where only the most extreme categories have labeled examples (i.e., $\mathcal{I}_k = \emptyset$ for intermediate categories), the optimization may collapse thresholds to eliminate unlabeled categories. Section \ref{sec:missing-categories} addresses this degeneracy through minimum threshold gap constraints.

\subsection{Mixed-Integer Programming Formulation}
\label{sec:mip}

We formulate the joint learning of integer scoring weights $\beta \in \mathbb{Z}^p$ and integer thresholds $\tau = (\tau_1, \ldots, \tau_{K-1}) \in \mathbb{Z}^{K-1}$ as a mixed-integer program that handles partial supervision and asymmetric costs. The MIP approach provides exact solutions and naturally incorporates both the integrality constraints and other discrete constraints essential for clinical deployment.

Table \ref{tab:variables} summarizes the decision variables and parameters in our formulation. The key insight is to use assignment variables $z_{ik}$ to determine the category assignment for each instance, with big-M constraints enforcing that assigned categories respect the threshold boundaries.

The complete MIP formulation is given by:
\begin{subequations}
\label{eq:mip}
\begin{align}
\min_{\beta, \tau, z} \quad & \sum_{i \in \mathcal{I}} w_i \sum_{k=1}^K \ell(k, k_i^*) \cdot z_{ik} \label{eq:mip-obj1} \\
\text{subject to} \quad & \sum_{k=1}^K z_{ik} = 1, \quad \forall i \in \mathcal{I} \label{eq:mip-assign1} \\ 
& \tau_k \leq \tau_{k+1} - \delta, \quad \forall k \in \{1,\ldots,K-2\} \label{eq:mip-mono} \\
& s_i = \beta^T x_i \quad 
\forall i \in \mathcal{I} \label{eq:score} \\
& M(z_{ik}-1) + \tau_{k-1} + \varepsilon \leq s_i, \notag \\
& \qquad \forall i \in \mathcal{I}, \, k \in \{1,\ldots,K\} \label{eq:mip-link1} \\
& s_i \leq M(1-z_{ik}) +\tau_k, \notag \\
& \qquad \forall i \in \mathcal{I}, \, k \in \{1,\ldots,K\} \label{eq:mip-link2} \\
& \beta \in \Omega \cap \mathbb{Z}^p \label{eq:governance}\\
& z \in \Theta \cap \{0,1\}^{K|\mathcal{I}|} \label{eq:performance} \\
& \tau \in \mathbb{Z}^{K-1} \label{eq:thresh-domain}
\end{align}
\end{subequations}

\begin{table}[tbh]
\centering
\caption{Decision variables and parameters in the MIP formulation}
\label{tab:variables}
\footnotesize
\setlength{\tabcolsep}{3pt}
\begin{tabular}{llp{3.5cm}}
\toprule
Symbol & Domain & Description \\
\midrule
$\beta$ & $\mathbb{Z}^p$ & Scoring weights \\
$\tau_k$ & $\mathbb{Z}$ & Threshold between categories $k$ and $k+1$ \\
$s_i$ & $\mathbb{Z}$ & Score for instance $i$: $s_i = \beta^\top x_i$ \\
$z_{ik}$ & $\{0,1\}$ & Assignment of instance $i$ to category $k$ \\
\midrule
$M$ & $\R_+$ & Big-M constant \\
$\varepsilon$ & $\R_+$ & Margin parameter \\
$\delta$ & $\mathbb{Z}_+$ & Min. gap between thresholds \\
\bottomrule
\end{tabular}
\end{table}

Equation \eqref{eq:mip-assign1} ensures each instance is assigned to exactly one category. Constraint \eqref{eq:mip-mono} maintains threshold ordering with minimum separation $\delta > 0$ to prevent degeneracy, as detailed in Section \ref{sec:missing-categories}. Constraints \eqref{eq:mip-link1}--\eqref{eq:mip-link2} implement the big-M formulation enforcing threshold bounds to assigned categories: if $z_{ik} = 1$, then $\tau_{k-1} + \varepsilon \leq s_i \leq \tau_k$. The margin $\varepsilon > 0$ ensures numerical stability and prevents ambiguity when scores lie exactly on a boundary. The domain constraints in \eqref{eq:governance} and \eqref{eq:performance} encapsulate governance and performance constraints on $\beta$ and $z$, respectively, that are described in Sections \ref{sec:interpretability} and \ref{sec:performance}.

We use an asymmetric ordinal loss function $\ell(k, k^*)$ in objective \eqref{eq:mip-obj1} that captures the differential consequences of misclassification:
\begin{equation}
\label{eq:ordinal-loss}
\ell(k, k^*) = \begin{cases}
\omega_{\text{under}} \cdot |k - k^*|^q & \text{if } k < k^* \\
\omega_{\text{over}} \cdot |k - k^*|^q & \text{if } k > k^* \\
0 & \text{if } k = k^*
\end{cases}, 
\end{equation}
where $\omega_{\text{under}} > \omega_{\text{over}}$ reflects that under-triage typically carries more severe consequences than over-triage in healthcare settings. The exponent $q \geq 1$ controls the growth rate: $q = 1$ yields linear penalty growth with ordinal distance, while $q = 2$ imposes quadratic penalties for severe misclassifications. The specific ratio $\omega_{\text{under}}/\omega_{\text{over}}$ should be determined based on the clinical context, weighing the relative costs of missed adverse events versus unnecessary interventions.

The big-M constraints in \eqref{eq:mip-link1}--\eqref{eq:mip-link2} require careful selection of $M$ to ensure correctness while maintaining numerical stability. We set $M = 2 \cdot \max_i \|x_i\| \cdot B_\beta + B_\tau$ where $B_\beta$ is an upper bound on $\|\beta\|$ and $B_\tau$ bounds the threshold range. These bounds can be derived from domain knowledge (e.g., reasonable score ranges in clinical tools) or data-driven estimates. The MIP has $O(nK)$ binary variables and $O(nK)$ constraints, making it tractable for moderate-sized clinical datasets (thousands of patients) using modern solvers. For larger problems, we employ the CSO warm-start strategy described in Section \ref{sec:two-phase}.

\subsection{Handling Missing Category Labels}
\label{sec:missing-categories}

In the extreme partial supervision setting, where certain categories lack any labeled examples, the optimization faces a fundamental challenge: without training instances to anchor middle categories, the objective may collapse thresholds to eliminate these categories entirely. While this maximizes performance on labeled extremes, it defeats the clinical purpose of having gradated risk levels that guide differential interventions. Consider the common scenario where only extreme cases can be reliably labeled: $k^*_i \in \{1, K\}$ for all training instances. An optimizer focused solely on separating these extremes might set $\tau_1 \approx \tau_2 \approx \cdots \approx \tau_{K-1}$, effectively collapsing all middle categories. This would achieve optimal separation on training data but produce a binary classifier rather than the desired ordinal tool.

Therefore, to maintain clinically meaningful risk stratification without fabricating labels for unlabeled categories, we impose minimum gap constraints per \eqref{eq:mip-mono}, where $\delta > 0$ ensures all categories maintain meaningful width. This parameter can be set through several approaches. In the case where we are optimizing an existing tool, we can utilize the existing tool's threshold gaps. For example, if the incumbent assessment tool has a gap of 8 points between low-risk and high-risk thresholds, we set $\delta = 8$. On a held-out set with known middle-category examples, we can select $\delta$ via cross-validation to maximize classification performance. Alternatively, we can apply clinical reasoning to set meaningful score differences (e.g., requiring at least 2--3 risk factors to differ between Low and High categories). For $K > 3$ categories, we can impose similar constraints on consecutive gaps or the total range, depending on which categories lack supervision.

\subsection{Governance Constraints}
\label{sec:interpretability}

The feasible set $\Omega$ in \eqref{eq:governance} encodes interpretability and deployment requirements essential for clinical adoption. These constraints ensure that optimized models remain transparent, clinically valid, and aligned with existing practice. Clinical knowledge about risk factor directionality can be encoded through sign restrictions on the weights. For factors known to increase risk, we enforce $\beta_j \geq 0$, while we enforce $\beta_j \leq 0$ for protective factors. This prevents counterintuitive weights that would undermine clinical trust, $\beta_j \geq 0 \quad \forall j \in \mathcal{J}_{\text{risk}},$ and $ \beta_j \leq 0 \quad \forall j \in \mathcal{J}_{\text{protective}}.$

To limit cognitive load during manual assessment, we can bound the number of active features through sparsity constraints. Clinical tools typically use 5--10 items for practical usability. Introducing binary variables $u_j$ to indicate whether feature $j$ is active, we enforce:
\begin{equation}
\label{eq:sparsity}
\sum_{j=1}^p u_j \leq s_{\max}, \quad -Mu_j \leq \beta_j \leq Mu_j, \quad u_j \in \{0,1\},
\end{equation}
where $s_{\max}$ bounds the total number of features with non-zero weights.

Clinical adoption is often facilitated by limiting changes from incumbent tools, as clinicians are more likely to accept refinements rather than wholesale replacements of existing practice. Let $\beta^{(0)}$ represent the incumbent weights. We can either enforce hard limits on individual changes or add a soft penalty to the objective:
\begin{equation}
\label{eq:minimal-edit}
|\beta_j - \beta_j^{(0)}| \leq \Delta_j \quad \text{or penalize} \quad \rho \sum_{j=1}^p |\beta_j - \beta_j^{(0)}|.
\end{equation}
The first form constrains each weight individually, while the second adds an $L_1$ penalty that encourages overall similarity to the incumbent model.

When features are inherently ordinal, monotonicity constraints can enforce the proper ordinal relationship. For example, the risk coefficient for taking two high fall risk medications should be at least as large as the coefficient for taking one high fall risk medication. Let $\mathcal{G}$ denote the set of feature pairs that should be ordered: $\beta_j \leq \beta_{j'} \quad \forall (j, j') \in \mathcal{G}.$

All of these interpretability constraints can be made to be linear and thus integrated directly into the MIP formulation, allowing optimization solvers to efficiently navigate the constrained solution space while maintaining clinical validity. The specific constraints deployed depend on the clinical context. Sign restrictions are nearly universal, sparsity is important when manual computation is required, minimal modification eases adoption of updated tools, and ordering ensures consistency and robustness against training data noise when features measure related constructs.

\subsection{Performance Constraints}
\label{sec:performance}

Performance constraints, encoded via the domain constraints in \eqref{eq:performance}, ensure the optimized model meets minimum safety and efficacy requirements. Using the assignment variables $z_{ik}$ from the MIP formulation, we can directly constrain classification error rates and bound operational metrics like sensitivity and specificity. For false positive rate control, we can limit the fraction of low-risk patients misclassified into higher risk categories. For example, to ensure that no more than $\xi_{\text{FPR}}$ fraction of truly low-risk patients are classified as high-risk:
\begin{equation}
\label{eq:fpr-constraint}
\frac{\sum_{i: k_i^* = 1} z_{iK}}{|\{i: k_i^* = 1\}|} \leq \xi_{\text{FPR}}.
\end{equation}

Similarly, for false negative rate control, we can limit the fraction of high-risk patients misclassified as low-risk. To ensure sensitivity of at least $1 - \xi_{\text{FNR}}$ for high-risk detection:
\begin{equation}
\label{eq:fnr-constraint}
\frac{\sum_{i: k_i^* = K} z_{i1}}{|\{i: k_i^* = K\}|} \leq \xi_{\text{FNR}}.
\end{equation}

More generally, for any true category $k^*$ and predicted category $k$, we can impose upper or lower bounds on misclassification rates:
\begin{equation}
\label{eq:general-performance}
\sum_{i \in \mathcal{I}_{k^*}} z_{ik} \leq u_{k^*,k} \cdot |\mathcal{I}_{k^*}| \quad \text{or} \quad \geq l_{k^*,k} \cdot |\mathcal{I}_{k^*}|,
\end{equation}
where $\mathcal{I}_{k^*} = \{i: \mathcal{S}_i = \{k^*\}\}$ denotes instances with known true label $k^*$, and $u_{k^*,k}$ (or $l_{k^*,k}$) specifies the maximum (or minimum) acceptable rate.

These constraints are linear in the assignment variables and integrate directly into the MIP formulation \eqref{eq:mip}. They are particularly valuable for ensuring that optimization does not sacrifice critical safety metrics (e.g., sensitivity for high-risk patients) in favor of overall accuracy. In practice, performance constraints can be derived from incumbent tool performance (ensuring the new model does not regress on key metrics) or from external regulatory or quality requirements.

\subsection{Constrained Score Optimization (CSO) Relaxation}
\label{sec:cso}

The MIP formulation provides exact solutions but faces computational challenges for large datasets. We derive a convex relaxation that preserves the essential structures of ordinal relationships, partial supervision, and asymmetric costs while enabling efficient optimization via gradient-based methods. The MIP model uses binary variables to encode category assignments and threshold crossings. The CSO relaxation replaces these discrete decisions with continuous margin-based losses. An important insight is that the ordinal loss decomposes into a sum of boundary crossing penalties, which we can approximate smoothly.

Consider predicting category $k$ for an instance with true category $k^*$. The ordinal loss $|k - k^*|$ equals the number of thresholds between $k$ and $k^*$. We can rewrite this as:
\begin{itemize}[leftmargin=*]
    \item If $k < k^*$: penalty for failing to cross thresholds $\tau_k, \tau_{k+1}, \ldots, \tau_{k^*-1}$.
    \item If $k > k^*$: penalty for incorrectly crossing thresholds $\tau_{k^*}, \tau_{k^*+1}, \ldots, \tau_{k-1}$.
\end{itemize}
\noindent This decomposition motivates a margin-based formulation where we penalize violations of desired threshold crossings.

To create a smooth approximation, we replace discrete boundary indicators with smooth softplus penalties. For each instance-threshold pair $(i, k)$, we define 
\begin{align}
\label{eq:softplus-detailed}
\phi_{i,k}^{-} &= \log\left(1 + \exp(\alpha^-(\tau_k - s_i))\right) \\
&\approx \begin{cases}
0 & \text{if } s_i \gg \tau_k \\
\alpha^-(\tau_k - s_i) & \text{if } s_i \ll \tau_k \\
\log(1+\exp(\alpha^-)) & \text{if } s_i = \tau_k
\end{cases}. \nonumber
\end{align}

\noindent Similarly, $\phi_{i,k}^{+} = \log(1 + \exp(\alpha^+(s_i - \tau_k)))$ penalizes exceeding threshold $k$ when we should not. The temperature parameters $\alpha^-$ and $\alpha^+$ control the steepness of the linear penalty for the score falling on the wrong side of the threshold.

To derive the CSO objective for instance $i$ with category label $k_i^*$, we define the CSO loss function as:
\begin{equation}
\label{eq:cso-loss-detailed}
\mathcal{L}_i(\beta, \tau) = \min_{k \in \Sset_i}  \sum_{j=1}^{k_i^*-1} \lambda_j^{+} \phi_{i,j}^{+} + \sum_{j=k_i^*}^{K-1} \lambda_j^{-} \phi_{i,j}^{-},
\end{equation}
where the weights $\lambda_j^{+}, \lambda_j^{-} > 0$ encode the asymmetric importance of each boundary. The full CSO formulation can be written as follows:
\begin{subequations}
\label{eq:cso-program-detailed}
\begin{align}
\min_{\beta, \tau} \quad & \sum_{i=1}^n w_i \mathcal{L}_i(\beta, \tau) + \mu R(\beta) \\
\text{subject to} \quad & \tau_1 \leq \tau_2 \leq \cdots \leq \tau_{K-1} \\
& \tau_{k+1} - \tau_k \geq \delta, \quad \forall k \in \{1, \ldots, K-2\} \\
& \beta \in \Omega
\end{align}
\end{subequations}
where $R(\beta)$ is an optional regularizer (e.g., $\|\beta\|_2^2$ for ridge, $\|\beta\|_1$ for lasso via proximal methods), and $\Omega$ encodes the same governance constraints as the MIP.

The CSO objective is smooth and thus differentiable everywhere with respect to both $\beta$ and $\tau$. Furthermore, the CSO objective \eqref{eq:cso-program-detailed} is convex in $(\beta, \tau)$ jointly because the softplus functions $\phi_{i,k}^{\pm}$ are convex. The weighted sum in \eqref{eq:cso-loss-detailed} preserves convexity. The differentiability and convexity of the objective, along with the convexity of the feasible set formed by the constraints allows for efficient solution of CSO using commercial nonlinear solvers.

\subsection{Two-Phase Optimization: CSO Warm-Start with MIP Refinement}
\label{sec:two-phase}

The CSO relaxation can provide a high-quality warm-start solution that significantly accelerates the subsequent MIP solve. The major challenge with implementing this warm-start approach is ensuring MIP feasibility of the warm-start solution. Simple rounding can be applied to adhere to integrality constraints for the risk model coefficients. Any continuous linear governance constraints from the MIP can be added to the CSO formulation without compromising differentiability and convexity. Performance constraints, on the other hand, cannot be directly added to CSO due to their reliance on the MIP binary variables. However, hyperparameter tuning of the CSO $\alpha$ and $\lambda$ values can be used to find solutions which satisfy these constraints.

This hybrid approach of using CSO and MIP combines the computational efficiency of convex optimization with the exactness of mixed-integer programming. The warm-start reduces solve time and helps avoid poor local optima by providing the MIP solver with a good initial feasible region.

\section{Experiments}
\label{sec:experiments}

The Johns Hopkins Fall Risk Assessment Tool provides an ideal test case for our framework as it exemplifies all the key challenges we address: labels available only for extreme categories due to intervention-censored outcomes, asymmetric costs where missing a fall is more harmful than over-treatment, and the need for interpretable integer-valued scores that clinicians can compute manually. In mapping JHFRAT to our general formulation, the feature vector $x_i \in \{0,1 \}^{17}$ represents the 17 binary risk factors across seven clinical categories (age, fall history, mobility, etc.), where $x_{ij} = 1$ indicates the presence of risk factor $j$ for patient $i$. The scoring function $s(x_i) = \beta^\top x_i$ computes the total JHFRAT score, where $\beta \in \mathbb{Z}^{17}_+$ contains the integer point values for each risk factor.

We evaluate our framework on optimizing the Johns Hopkins Fall Risk Assessment Tool using electronic health record data from three Johns Hopkins Health System hospitals: Johns Hopkins Hospital, Johns Hopkins Bayview Medical Center, and Howard County Medical Center. Our dataset comprises 54,209 hospital admissions between March 28, 2022 and October 27, 2023, with complete JHFRAT assessments, intervention records, and fall events. The JHFRAT assessment consists of 17 non-zero coefficient risk factors across seven categories: age, bladder/bowel elimination, cognition, fall history, patient care equipment, medications, and mobility. On average, patients are assessed twice daily with JHFRAT, and our dataset has an average of 1.87 JHFRAT records per patient day. This study was approved by the Johns Hopkins University Institutional Review Board: IRB00395329, Functional measures in the hospital and patient outcomes.

\subsection{Dataset Construction and Partial Labels}

A total of 498 hospital encounters in the dataset include at least one fall event, constituting 0.92\% of encounters and equating to 1.07 falls per 1,000 patient-days. The majority of patients in the data receive one or more fall-prevention interventions during hospitalization. Of the 31 possible fall-prevention interventions, 13 were identified by a team of clinicians as ``targeted'' for being resource-intensive and capable of meaningfully altering fall risk. Such targeted interventions include the use of bed exit alarms, increased patient rounding, and constant monitoring. We assume that these targeted interventions, due to their intensive nature and established efficacy, can meaningfully obscure underlying fall risk. In contrast, we do not consider general interventions, like educating patients to wait/call for assistance prior to mobilizing, to be risk-obscuring as they are universally applied and have minimal direct impact on fall occurrence. The partial supervision challenge manifests clearly in this setting: among 54,209 hospital encounters, the vast majority (80.7\%) received targeted fall prevention interventions that obscure their true risk. The rarity of fall events, combined with widespread use of preventive interventions, creates the selective labeling challenge central to our approach.

For patients who experienced any falls, we consider only the JHFRAT assessments prior to the first fall event. To obtain a medically representative sample for the study cohort, we include patients who fell between the 2nd and 21st days of the encounter, or whose overall length of stay was between 2 and 21 full days. We exclude all encounters with fewer than two JHFRAT assessments during the included encounter time. After applying this filtering, the study cohort includes 46,991 patient encounters.

\begin{figure}[tbh]
\centering
\includegraphics[width=0.45\textwidth]{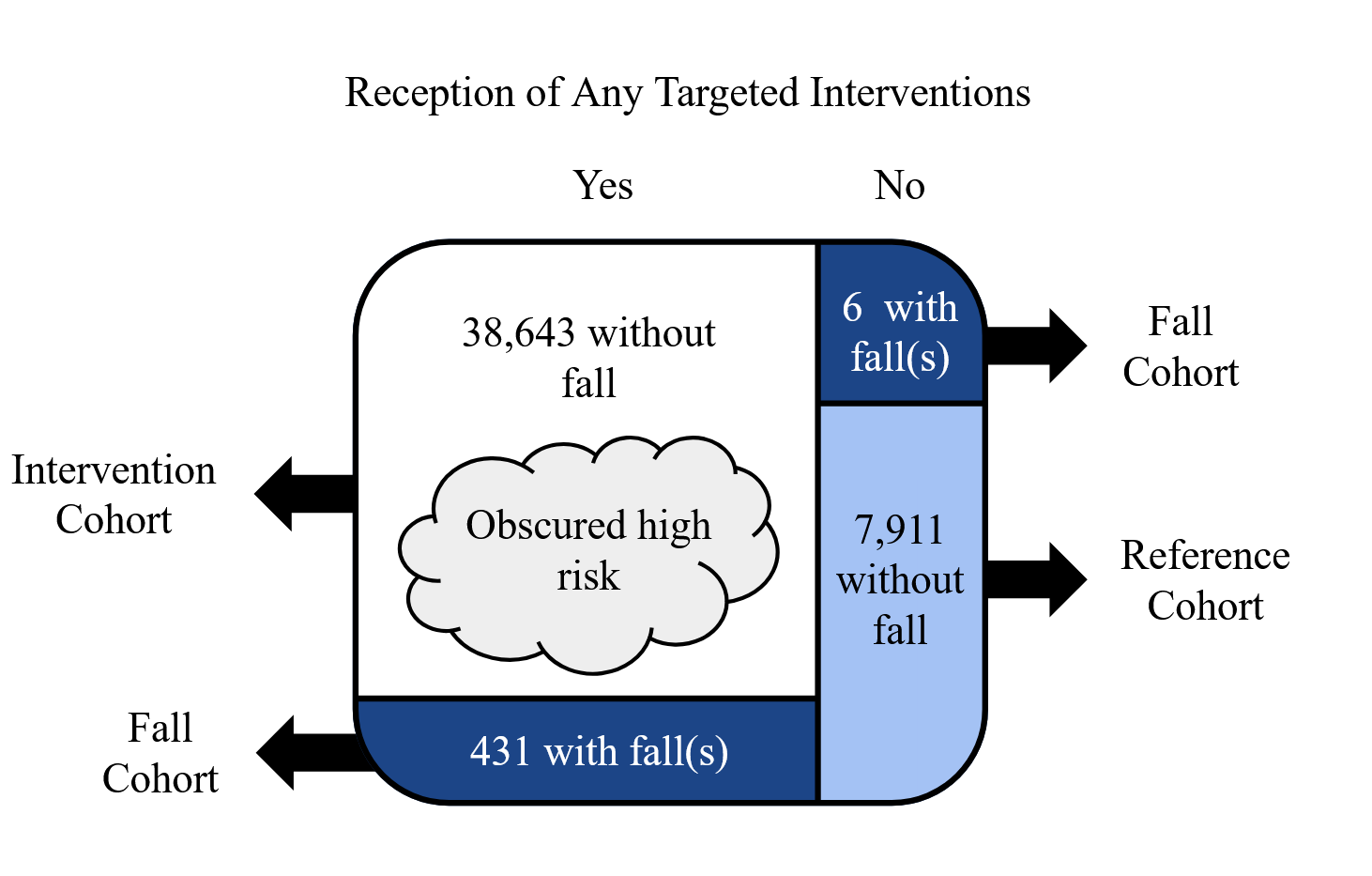}
\caption{Stratification of inpatient admissions based on fall events and targeted interventions. The intervention cohort was excluded from primary optimization analyses but retained for post-optimization validation.}
\label{fig:cohort_division}
\end{figure}

Figure \ref{fig:cohort_division} illustrates how we partition the 46,991 filtered patient encounters into three cohorts based on fall events and targeted intervention receipt. The key insight is that intervention receipt creates asymmetric information about true risk: while falls indicate high risk regardless of interventions (since interventions failed to prevent them), the absence of falls can only be construed as an indicator of true low risk if the patient received no targeted interventions.

We partition patient encounters into three cohorts with distinct labeling strategies:
\begin{itemize}[leftmargin=*]
\item \textbf{Fall cohort:} All encounters with at least one fall event, regardless of targeted interventions applied. These encounters are labeled high-risk.
\item \textbf{Reference cohort:} Encounters with no targeted interventions throughout the encounter, and no fall events. These encounters are labeled low-risk. 
\item \textbf{Intervention cohort:} Encounters with at least one targeted intervention during the encounter, and no fall events. These encounters are considered to have uncertain risk, and are not labeled for the optimization. 
\end{itemize}

The extreme imbalance in cohort sizes underscores why standard supervised learning approaches fail in this setting. Simply assigning the 38,643 intervention cohort patients to a risk category based on their JHFRAT scores or fall outcomes would introduce massive label noise, as we would be guessing the counterfactual (what would have happened without interventions) for the majority of our training data. Our partial supervision approach instead acknowledges this uncertainty explicitly, training only on the 8,348 patients (17.8\% of included encounters) for whom we have confident labels while reserving the intervention cohort for post-optimization validation to assess how the optimized models generalize to the ambiguous majority.

\subsection{MIP Models}

The JHFRAT scale is divided, per current practice, into three risk categories: Low-Risk ($k=1$) for scores less than 6, Moderate-Risk ($k=2$) for scores 6-13, and High-Risk ($k=3$) for scores greater than 13. We apply the corresponding true labels to the labeled training data: $k^* = 1$ for the low-risk patients in the reference cohort and $k^* = 3$ for the high-risk patients in the fall cohort. While our general framework jointly optimizes both scoring weights $\beta$ and thresholds $\tau$, the JHFRAT application presents unique deployment constraints that motivate a modified approach. To maintain compatibility with existing clinical workflows, electronic health record systems, and staff training, we fix the thresholds at their current values ($\tau_1 = 6, \tau_2 = 13$) and focus optimization on the scoring weights $\beta$ alone. This decision reflects a common practical scenario where healthcare systems require incremental refinements to existing tools rather than complete redesigns, as changing risk thresholds would necessitate updates to intervention protocols, clinical guidelines, and quality metrics across the entire health system.

We note that this fixed-threshold approach, while more restrictive than full joint optimization, offers several practical advantages: (1) it ensures the optimized tool remains immediately deployable without systemic changes to care protocols, (2) it maintains comparability with historical JHFRAT data and quality metrics, and (3) it reduces the optimization complexity, improving solution stability and interpretability. The resulting problem focuses on finding optimal integer weights $\beta \in \mathbb{Z}^{17}_+$ that best separate the extreme risk groups within the established scoring scale, effectively treating this as a constrained parameter refinement problem rather than a full model redesign. The complete JHFRAT-optimizing MIP formulation with these fixed thresholds is given as follows:
\begin{subequations}
\label{eq:jhfrat-mip}
\begin{align}
\min_{\beta, \tau, z} \quad & \sum_{i=1}^n w_i \sum_{k=1}^3 \ell(k, k_i^*) \cdot z_{ik} \label{eq:jhfrat-mip-obj} \\
\text{subject to} \quad & \sum_{k=1}^3 z_{ik} = 1, \quad \forall i \label{eq:mip-assign2} \\
& \text{(1d) - (1h)} \notag \\
& 0 \leq \beta_j \leq 13 \quad \forall j \label{eq:beta-range} \\
& \beta_i \leq \beta_j \quad \forall (i,j) \in P \label{eq:beta-ordering}\\
& \frac{\sum_{i \in [n] : k_i^* = 3} z_{i1}}{|\{i:k^*_i = 3 \}|} \leq \xi_{sFNR} \label{eq:FNcap-severe}\\
& \frac{\sum_{i \in [n] : k_i^* = 3} z_{i2}}{|\{i:k^*_i = 3 \}|} \leq \xi_{mFNR} \label{eq:FNcap-moderate} \\
& \frac{\sum_{i \in [n] : k_i^* = 1} z_{i3}}{|\{i:k^*_i = 1 \}|} \leq \xi_{sFPR} \label{eq:FPcap-severe}\\
& \frac{\sum_{i \in [n] : k_i^* = 1} z_{i2}}{|\{i:k^*_i = 1 \}|} \leq \xi_{mFPR} \label{eq:FPcap-moderate}\\
& \beta_j \in \mathbb{Z}, \quad z_{ik} \in \{0,1\} \label{eq:mip-domain}
\end{align}
\end{subequations}

The objective function \eqref{eq:jhfrat-mip-obj} exactly follows from \eqref{eq:mip-obj1} by optimizing over all $n$ training samples and 3 JHFRAT categories. The definition for $\ell(k, k_i^*)$ follows from \eqref{eq:ordinal-loss}, and we set $\omega_{under} = 3$ and $\omega_{over} = 1$ so that the ordinal loss for under-triage is three times that of over-triage. The sample weights $w_i$ are derived from the number of samples in each class such that the total contribution to the objective function is equal between the imbalanced low and high-risk classes: $w_i = \frac{n}{|\{j : k_j^* = k_i^*\}|}.$

We apply three types of governance constraints in our MIP formulations for JHFRAT. Firstly, we apply a uniform lower and upper bound on all model coefficients via \eqref{eq:beta-range}. Since we only include the risk factors from JHFRAT with positive coefficients, it is reasonable to impose the clinical assumption inherent in the lower bound of 0 that all of these factors can only contribute positively to high fall risk. Furthermore, we limit coefficients to the highest class's lower bound of 13 for model stability. Secondly, in \eqref{eq:beta-ordering} we impose monotonicity constraints for hierarchical risk factors to maintain clinical validity. In particular, the risk factors in the categories of age, medications, and patient care equipment are inherently hierarchical, and we let $P$ denote the set of pairs of risk factor indices corresponding to these pairwise orderings (i.e., the model coefficient for one high-fall risk drug is less than or equal to the coefficient for two or more high-fall risk drugs). Finally, in \eqref{eq:mip-domain} we impose integer constraints on the coefficients to maintain consistency with the format of JHFRAT and ease cognitive burden of manual assessment.

We impose four performance constraints using the number of false positives and false negatives from the original JHFRAT model. The number of High-Risk labeled encounters classified as Low-Risk \eqref{eq:FNcap-severe} and Moderate Risk \eqref{eq:FNcap-moderate} cannot exceed the rates of such ``severe" and ``moderate" false-negative classifications with the JHFRAT model, $\xi_{sFNR}$ and $\xi_{mFPR}$. Conversely, the number of Low-Risk labeled encounters classified as High-Risk \eqref{eq:FPcap-severe} and Moderate Risk \eqref{eq:FPcap-moderate} cannot exceed the rates of the severe and moderate false-positive classifications with the JHFRAT model, $\xi_{sFPR}$ and $\xi_{mFPR}$. These performance constraints ensure that improvements achieved by the MIP model do not come at the expense of worse performance for one class or the other.

\subsection{Experimental Setup}

We split the encounters from the combined fall and reference cohorts into an 80\% training set, and a 20\% test set, stratified by risk label and current JHFRAT category. We retain the intervention cohort separately as an exogenous dataset for concordance analysis post-optimization. We first train a series of CSO models via a hyperparameter gridsearch over the values of $\alpha^-$, $\alpha^+ \in \{0.5, 1, 2, 4 \}$ and $\lambda_1 \in \{0.2, 0.5, 0.8\}$ (such that $\lambda_1 = \lambda_1^- = \lambda_1^+$ and $\lambda_2 = 1- \lambda_1 = \lambda_2^- = \lambda_2^+$). For each set of hyperparameters, we perform 5-fold cross-validation over the training set and average the coefficients across the cross-validation folds. The set of candidate warm-start solutions is the sets of these averaged coefficients, rounded to the nearest integer.

The MIP model is trained on the entire training set, utilizing the candidate warm-start solution that has the best MIP objective value while satisfying all MIP constraints. If no candidate warm-start solutions are feasible, we warm-start with the original JHFRAT coefficients. The CSO models are solved with MOSEK via cvxpy, and all MIP variations are solved with Gurobi 12.0.3.

\subsection{Baselines and Evaluation Metrics}

We compare the outcomes of our approach against the original JHFRAT model with expert-derived weights. Furthermore, we compare the CSO performance with that of the MIP models in order to evaluate the added value of the MIP portion of the optimization pipeline. We utilize the following metrics to compare model performance:

\subsubsection{Classification Accuracy} 
An encounter data sample is considered correctly classified only if it is exactly classified into its correct low-risk or high-risk category. A data sample is considered mis-classified with ``moderate error'' if it is classified into the moderate-risk category, and mis-classified with ``severe error'' if it is mis-classified into the opposite category: a low-risk labeled sample classified into the high-risk category and vice-versa. We define \textit{tight accuracy} as the portion of encounters correctly classified, and define \textit{loose accuracy} to additionally include all instances of moderate error as correctly classified.

\subsubsection{High-Risk Precision and Recall} 
We rely on the high-risk precision and recall to account for the imbalance of the dataset and the importance of correctly identifying high-risk patients in this setting. For these binary-classification-based metrics, moderate and low-risk classifications are grouped together.

\subsubsection{Binary Classification Performance Measures} 
We utilize the distribution of scores for the optimized models to generate receiver operating characteristic (ROC) and precision-recall curves. We then utilize the area under each curve, AUROC and AUPRC, respectively, as model performance metrics.

\section{Results}
\label{sec:results}

The optimal solution to the MIP defined in \eqref{eq:jhfrat-mip} is featured in Table \ref{tab:coefficients}, in comparison with the current JHFRAT scoring coefficients. The optimal feasible CSO warm-start coefficients for this MIP (hyperparameters $\alpha^- = 0.5, \alpha^+ = 1, \lambda_1^-=\lambda_1^+=0.8, \lambda_2^-=\lambda_2^+=0.2$) are also included. Both the MIP and CSO warm-start have total coefficient sums greater than the current JHFRAT, promoting higher possible total scores. Cognition and elimination risk factors feature far more prominently in the MIP model than in the current JHFRAT. Conversely, all patient care equipment coefficients are zero in the optimal MIP model.

\begin{table}[ht]
\centering
\scriptsize
\caption{Risk factor coefficients across baseline and optimized models}
\label{tab:coefficients}
\begin{tabular}{ ll|ccc } 
\toprule
Category & Risk Factor & JHFRAT & CSO & MIP \\ 
\hline
\multirow{3}{2em}{Age} & 60-69 years & 1 & 1 & 0 \\
& 70-79 years & 2 & 1 & 2 \\
& $\geq$ 80 years & 3 & 2 & 2 \\
\hline
\multirow{2}{2em}{Elimination} & Incontinence  & 2 & 6 & 6 \\
& Urgency/frequency  & 2 & 3 & 7 \\
\hline
\multirow{3}{2em}{Cognition} & Altered awareness  & 1 & 6 & 13 \\
& Impulsive  & 2 & 12 & 12 \\
& Lack of understanding  & 4 & 3 & 4 \\
\hline
\multirow{3}{2em}{Equipment} & 1 present  & 1 & 1 & 0 \\
& 2 present & 2 & 1 & 0 \\
& 3+ present & 3 & 1 & 0 \\
\hline
Fall Hist. & Fall within 6 months & 5 & 2 & 1 \\
\hline 
\multirow{3}{2em}{Medication} & 1 high fall risk drug & 3 & 3 & 2 \\
& 2+ high fall risk drugs & 5 & 4 & 2 \\
& Sedation procedure & 7 & 4 & 2 \\
\hline
\multirow{3}{2em}{Mobility} & Requires assistance & 2 & 4 & 7 \\
& Unsteady gait & 2 & 4 & 2 \\
& Visual/auditory impairment & 2 & 6 & 3 \\
\hline
& \textbf{Total Coeff. Sum} & \textbf{49} & \textbf{64} & \textbf{65} \\
\bottomrule
\end{tabular}
\end{table}

\begin{figure}[ht]
\centering
\includegraphics[width=0.48\textwidth]{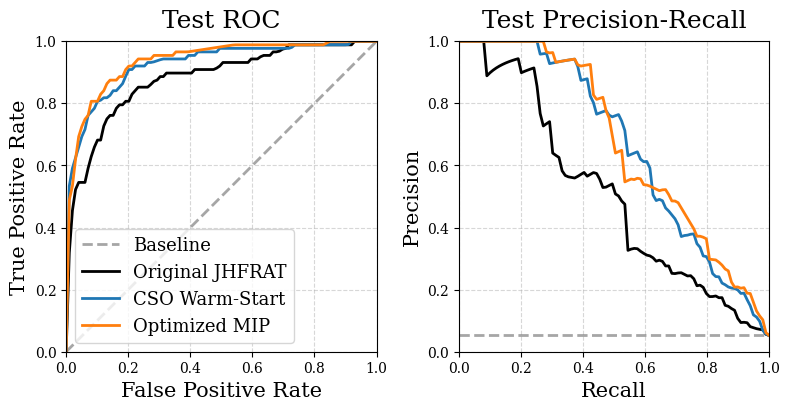}
\caption{Receiver Operating Characteristic (ROC) and Precision-Recall performance curves for JHFRAT and optimized models on testing dataset. Both the CSO warm-start and optimized MIP improve over the original JHFRAT.}
\label{fig:testing-curves}
\end{figure}

Table \ref{tab:traintest-accuracy} details the model categorization accuracy per labeled class for each model. Both optimized models improve correct classification on the training data for both low-risk and high-risk patients while reducing or maintaining the rates of moderate and severe error, as designed. For the training data, correct low-risk classification increases from 71\% (original JHFRAT) to 83\% (MIP), and correct high-risk classification increases from 30\% to 46\%. This improvement persists in the testing data, with an increases to 84\% and 48\% correct classification for low-risk and high-risk patients respectively, with the JHFRAT testing data rates remaining the same. Furthermore, the severe and moderate error rates for the models remain the same or lower than the JHFRAT rates for the testing data. The severe error rates remain comparable across models, likely because the severe error performance constraints (5e) and (5g) are in direct conflict with each other: there is a natural limit, to how much the model can reduce the severe false positives without increasing severe false negatives, and vice versa. For the training data, these constraints are explicitly enforced, but the effects of these constraints persist in the testing data, as evident by the consistency in severe error rates for testing data. The optimized models outperform JHFRAT in area under the precision-recall curve for the testing datasets, per Table \ref{tab:testing-metrics} and Figure \ref{fig:testing-curves}. The MIP's focus on categorizing around thresholds is reflected in its improved accuracy metrics over the CSO model, despite having similar AUC scores.

\begin{table}[ht]
\centering
\caption{Train and test data categorization comparison across original and optimized models}
\label{tab:traintest-accuracy}
\begin{tabular}{ ll | ccc } 
\toprule
 \multirow{2}{4em}{\textbf{Model}} & \multirow{2}{4em}{\textbf{Label}} & \multicolumn{3}{c}{\textbf{Categorization}}  \\ 
 &  & Correct & Mod. Err. & Severe Err. \\ \midrule
 \multicolumn{5}{c}{\textbf{Training Data}} \\ \hline
 \multirow{2}{6em}{Orig. JHFRAT} & Low-Risk & 4,519 (71\%) & 1,788 (28\%) & 58 (1\%)  \\  
 & High-Risk & 105 (30\%) & 202 (58\%) & $\bm{42 (12\%)}$ \\
 \hline
  \multirow{2}{6em}{CSO} & Low-Risk & 4,940 (78\%) & 1,377 (22\%) &  $\bm{48 (1\%)}$ \\  
 & High-Risk & 151 (43\%) & 156 (45\%) & $\bm{42 (12\%)}$  \\ 
 \hline
 \multirow{2}{6em}{Opt. MIP} & Low-Risk & $\bm{5,271 (83\%)}$ & $\bm{1,036 (16\%)}$ &  58 (1\%) \\  
 & High-Risk & $\bm{160 (46\%)}$ & $\bm{147 (42\%)}$ & $\bm{42 (12\%)}$  \\ \midrule
  \multicolumn{5}{c}{\textbf{Testing Data}} \\ \hline
   \multirow{2}{6em}{Orig. JHFRAT} & Low-Risk & 1,135 (71\%) & 441 (28\%) & 15 (1\%)  \\  
 & High-Risk & 26 (30\%) & 51 (58\%) & $\bm{11 (13\%)}$ \\
 \hline
   \multirow{2}{6em}{CSO} & Low-Risk & 1,279 (80\%) & 301 (19\%) &  $\bm{11 (1\%)}$ \\  
 & High-Risk & 39 (45\%) & 40 (45\%) & $\bm{9 (10\%)}$  \\ 
 \hline
 \multirow{2}{6em}{Opt. MIP} & Low-Risk & $\bm{1,342 (84\%)}$& $\bm{235 (15\%)}$ &  14 (1\%) \\  
 & High-Risk & $\bm{42 (48\%)}$ & $\bm{35 (40\%)}$ & $\bm{11 (13\%)}$  \\ 
 \bottomrule
\end{tabular}
\end{table}

\begin{table}[ht]
\centering
\caption{Performance metrics on test dataset}
\label{tab:testing-metrics}
\footnotesize
\setlength{\tabcolsep}{3pt}
\begin{tabular}{ l | cccccc } 
\toprule
 \textbf{Model} & \textbf{Tight Acc.} & \textbf{Loose Acc.} & \textbf{ROC} & \textbf{PRC} & \textbf{Prec.} & \textbf{Rec.}  \\
 \hline
Orig. JHFRAT & 0.69  & 0.98 & 0.88 & 0.51 & 0.63 & 0.29 \\
CSO & 0.79 & $\bm{0.99}$ & 0.93 & 0.66 & $\bm{0.78}$ & 0.44 \\
Opt. MIP & $\bm{0.82}$ & 0.98 & $\bm{0.94}$ & $\bm{0.67}$ & 0.75 & $\bm{0.48}$ \\
 \bottomrule
\end{tabular}
\end{table}

\begin{figure}[ht]
\centering
\includegraphics[width=0.48\textwidth]{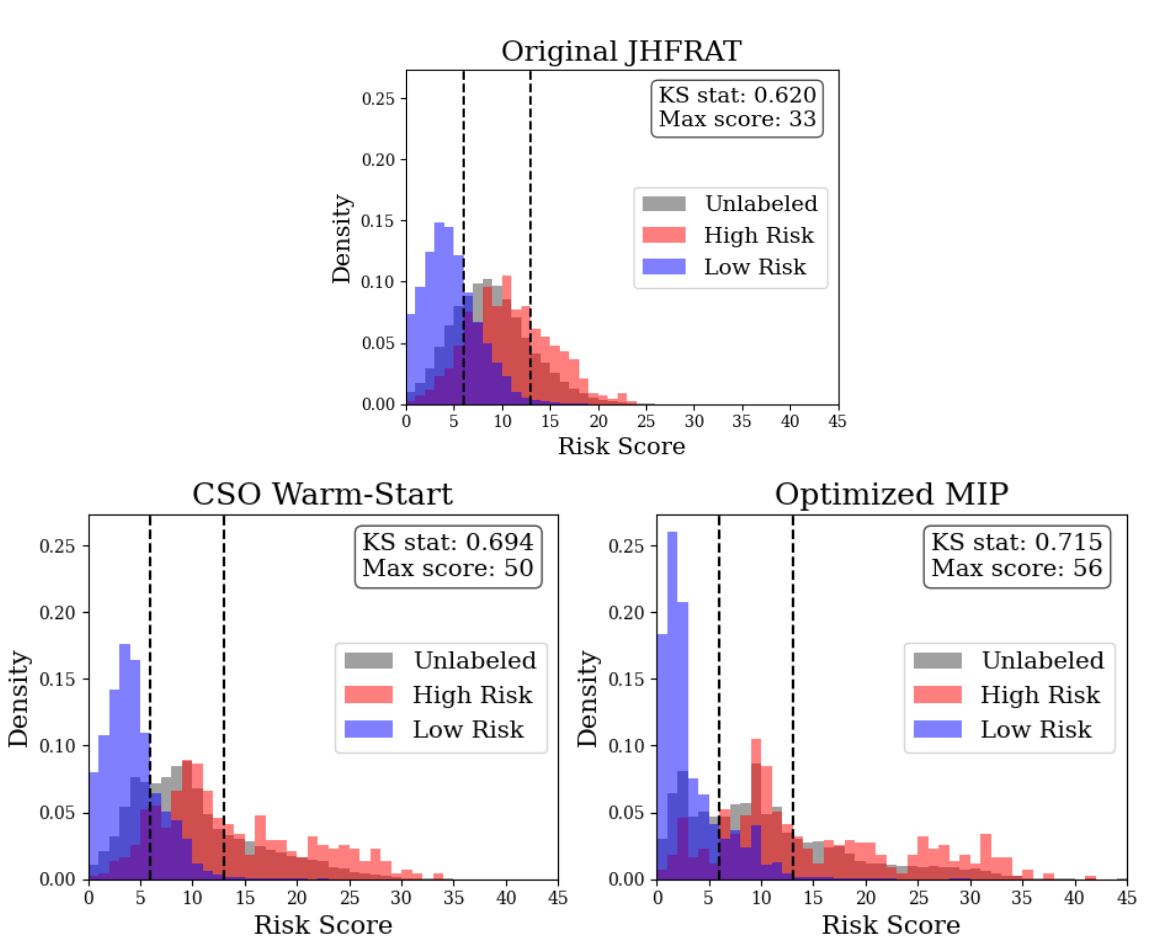}
\caption{Score distributions for all study data (test and train data together). Both CSO and MIP improve separation of high-risk and low-risk samples, at the expense of skewing high-risk and unlabeled samples to higher scores.}
\label{fig:score_distributions}
\end{figure}

Figure \ref{fig:score_distributions} reveals how the optimized model coefficients reshape the risk score distributions compared to the original JHFRAT. The Kolmogorov-Smirnov (KS) statistic quantifies the maximum vertical separation between the cumulative distribution functions of the Safe Low-Risk and Safe High-Risk cohorts, with larger values indicating better discriminative ability. The associated p-value tests the null hypothesis that the two risk groups come from the same distribution. The original JHFRAT, CSO warm-start and optimized MIP models all have highly significant p-values (all p < 0.001), confirming that all models successfully separate the risk groups, though with varying degrees of separation. The distribution of the unlabeled intervention cohort is included for reference.

The overlapping distributions in the original JHFRAT model (KS = 0.620) show why current practice struggles: many true high-risk patients score in the low-to-moderate range (under-triage), while some low-risk patients score high (over-triage). This overlap drives both missed falls and unnecessary interventions. Both the CSO warm-start and optimized MIP improve on the KS statistic, at the expense of skewing the high-risk and unlabeled distributions rightward and increasing the maximum total scores from 33 for the original JHFRAT to 50 and 55 for CSO and MIP, respectively.

\section{Discussion}
\label{sec:discussion}

The proposed framework makes several methodological contributions to the optimization of clinical risk scoring tools. First, the formulation addresses a common challenge in clinical settings where reliable labels exist only for extreme risk categories. When optimizing with only these extreme labels, the model may collapse intermediate thresholds to eliminate unlabeled middle categories, effectively producing a binary classifier rather than the desired ordinal tool. The framework prevents this degeneracy through minimum threshold gap constraints, maintaining clinically meaningful risk stratification without requiring labeled examples for every category. Second, the asymmetric distance-aware loss function directly encodes clinical priorities into the optimization objective. By allowing differential penalties for under-triage versus over-triage that scale with ordinal distance, the framework aligns the training signal with deployment consequences, which is preferable to learning symmetric costs and adjusting thresholds post-hoc. Third, the governance and performance constraints provide mechanisms for encoding domain knowledge and ensuring operational safety, from sign restrictions that maintain clinical face validity to false-positive and false-negative caps that guarantee minimum performance levels. Finally, the two-phase optimization pipeline combines the computational efficiency of convex optimization with the exactness of mixed-integer programming, enabling tractable solutions for moderate-sized clinical datasets.

The JHFRAT case study validates these methodological contributions empirically. Training only on the 8,348 patients with confident extreme labels while excluding the 38,643 with intervention-censored outcomes yields models that generalize well to held-out test data, with tight accuracy improving from 69\% to 82\% and high-risk recall nearly doubling from 29\% to 48\%. The asymmetric loss function (3:1 under-triage to over-triage penalty) successfully biases the optimization toward high-risk identification without sacrificing overall discrimination, as evidenced by the AUPRC improvement from 0.51 to 0.67. The two-phase pipeline demonstrates practical value: the CSO warm-start achieves 79\% tight accuracy on its own and provides feasible integer solutions that accelerate MIP convergence, while the MIP refinement yields additional gains in threshold-based classification. The similar AUC scores but improved accuracy metrics between CSO and MIP suggest that while both methods learn comparable score rankings, the MIP's discrete optimization produces coefficients better calibrated to the fixed threshold boundaries.

Beyond performance improvements, the optimized coefficients reveal clinically meaningful patterns that illustrate the value of data-driven calibration. Cognition-related factors, particularly altered awareness and impulsivity, receive substantially higher weights in the optimized model, increasing from 1--2 points to 12--13 points. Elimination-related factors similarly receive increased weights, aligning with research on bathroom-related ambulation as a high-risk period. Conversely, patient care equipment coefficients are reduced to zero, possibly reflecting confounding in the original design where equipment presence correlates with closer monitoring or serves as a proxy for illness severity captured elsewhere. These discoveries would not emerge from simply re-thresholding the original scores; they require the coefficient optimization that the framework enables.

Several limitations warrant acknowledgment. The use of encounter-averaged JHFRAT scores introduces non-integer feature values, which may contribute to the large coefficient magnitudes observed; future work could optimize directly on assessment-level data or impose tighter coefficient bounds. The complete exclusion of the intervention cohort, while methodologically principled, may limit generalizability to patients who would typically receive interventions. Incorporating partial labels based on the specific interventions applied or otherwise implementing multi-label formulations could address this. The fixed-threshold constraint, while practically motivated for deployment compatibility, prevents full exploitation of the joint optimization framework. Additionally, the widened score range (maximum scores increasing from 33 to 56) may affect clinician interpretation of within-category gradation, requiring communication and training for deployment.

The framework applies broadly to clinical risk stratification settings sharing the key structural features: ordinal categories from point-based scores, labels available only at extreme categories, and asymmetric misclassification costs. Examples include pressure injury risk (Braden Scale), venous thromboembolism risk (Caprini Score), and early warning scores for deterioration. Future extensions include non-linear scoring functions maintaining interpretability, temporal dynamics for time-varying risk, and federated optimization across institutions with heterogeneous populations.

\section{Conclusion}
\label{sec:conclusion}

This work presents a mixed-integer programming framework for optimizing clinical risk scoring tools in settings where reliable labels exist only for extreme risk categories. The framework addresses three challenges that impede standard supervised learning in clinical deployment: missing labels for intermediate categories that can cause threshold collapse, asymmetric misclassification costs that vary with ordinal distance, and interpretability requirements that mandate integer coefficients amenable to manual calculation. Minimum threshold gap constraints maintain clinically meaningful risk stratification without requiring labeled examples for every category, while the constrained score optimization relaxation provides an efficient warm-start strategy that accelerates MIP convergence.

Applied to the Johns Hopkins Fall Risk Assessment Tool, the optimized model improves high-risk recall from 29\% to 48\% and area under the precision-recall curve from 0.51 to 0.67, while maintaining the existing three-tier threshold structure for seamless integration with current clinical workflows. The optimized coefficients reveal that cognition and elimination factors are substantially more predictive than their original weights suggest, while patient care equipment contributes minimal independent information. These insights may inform future tool revisions beyond the current optimization.

The framework's flexibility in accommodating governance constraints, performance guarantees, and minimal-edit requirements positions it for broader application across clinical risk assessment domains where data-driven calibration of legacy tools is needed. Future extensions may incorporate time-varying risk dynamics, multi-site federated learning, and richer label structures that leverage intervention cohort information more directly.

\bibliographystyle{IEEEtran}
\bibliography{references}

@article{cao2020rank,
  title={Rank consistent ordinal regression for neural networks with application to age estimation},
  author={Cao, Wenzhi and Mirjalili, Vahid and Raschka, Sebastian},
  journal={Pattern Recognition Letters},
  volume={140},
  pages={325--331},
  year={2020},
  publisher={Elsevier}
}

@article{chu2007support,
  title={Support vector ordinal regression},
  author={Chu, Wei and Keerthi, S Sathiya},
  journal={Neural computation},
  volume={19},
  number={3},
  pages={792--815},
  year={2007},
  publisher={MIT Press One Rogers Street, Cambridge, MA 02142-1209, USA journals-info~…}
}

@article{cour2011learning,
  title={Learning from partial labels},
  author={Cour, Timothee and Sapp, Ben and Taskar, Ben},
  journal={The Journal of Machine Learning Research},
  volume={12},
  pages={1501--1536},
  year={2011},
  publisher={JMLR. org}
}

@inproceedings{wang2019partial,
  title={Partial Label Learning with Unlabeled Data.},
  author={Wang, Qian-Wei and Li, Yufeng and Zhou, Zhi-Hua and others},
  booktitle={IJCAI},
  pages={3755--3761},
  year={2019}
}

@article{feng2020provably,
  title={Provably consistent partial-label learning},
  author={Feng, Lei and Lv, Jiaqi and Han, Bo and Xu, Miao and Niu, Gang and Geng, Xin and An, Bo and Sugiyama, Masashi},
  journal={Advances in neural information processing systems},
  volume={33},
  pages={10948--10960},
  year={2020}
}

@article{goldstein2016opportunities,
  title={Opportunities and challenges in developing risk prediction models with electronic health records data: a systematic review},
  author={Goldstein, Benjamin A and Navar, Ann Marie and Pencina, Michael J and Ioannidis, John PA},
  journal={Journal of the American Medical Informatics Association: JAMIA},
  volume={24},
  number={1},
  pages={198},
  year={2016}
}

@article{rudin2019stop,
  title={Stop explaining black box machine learning models for high stakes decisions and use interpretable models instead},
  author={Rudin, Cynthia},
  journal={Nature machine intelligence},
  volume={1},
  number={5},
  pages={206--215},
  pages={206--215},
  year={2019},
  publisher={Nature Publishing Group UK London}
}

@inproceedings{lv2020progressive,
  title={Progressive identification of true labels for partial-label learning},
  author={Lv, Jiaqi and Xu, Miao and Feng, Lei and Niu, Gang and Geng, Xin and Sugiyama, Masashi},
  booktitle={international conference on machine learning},
  pages={6500--6510},
  year={2020},
  organization={PMLR}
}

@article{mccullagh1980regression,
  title={Regression models for ordinal data},
  author={McCullagh, Peter},
  journal={Journal of the Royal Statistical Society: Series B (Methodological)},
  volume={42},
  number={2},
  pages={109--127},
  year={1980},
  publisher={Wiley Online Library}
}

@inproceedings{niu2016ordinal,
  title={Ordinal regression with multiple output cnn for age estimation},
  author={Niu, Zhenxing and Zhou, Mo and Wang, Le and Gao, Xinbo and Hua, Gang},
  booktitle={Proceedings of the IEEE conference on computer vision and pattern recognition},
  pages={4920--4928},
  year={2016}
}

@article{oliver1997development,
  title={Development and evaluation of evidence based risk assessment tool (STRATIFY) to predict which elderly inpatients will fall: case-control and cohort studies},
  author={Oliver, David and Britton, M and Seed, P and Martin, FC and Hopper, AH},
  journal={Bmj},
  volume={315},
  number={7115},
  pages={1049--1053},
  year={1997},
  publisher={British Medical Journal Publishing Group}
}

@article{jette2014pac,
  title={AM-PAC “6-Clicks” functional assessment scores predict acute care hospital discharge destination},
  author={Jette, Diane U and Stilphen, Mary and Ranganathan, Vinoth K and Passek, Sandra D and Frost, Frederick S and Jette, Alan M},
  journal={Physical therapy},
  volume={94},
  number={9},
  pages={1252--1261},
  year={2014},
  publisher={Oxford University Press}
}

@inproceedings{caprini1991clinical,
  title={Clinical assessment of venous thromboembolic risk in surgical patients.},
  author={Caprini, JA and Arcelus, JI and Hasty, JH and Tamhane, AC and Fabrega, F},
  booktitle={Seminars in thrombosis and hemostasis},
  volume={17},
  pages={304--312},
  year={1991}
}

@article{bergstrom1987braden,
  title={The Braden Scale for predicting pressure sore risk},
  author={Bergstrom, Nancy and Braden, Barbara J and Laguzza, Antoinette and Holman, Victoria},
  journal={Nursing research},
  volume={36},
  number={4},
  pages={205--210},
  year={1987},
  publisher={LWW}
}

@article{poe2007johns,
  title={The Johns Hopkins fall risk assessment tool: postimplementation evaluation},
  author={Poe, Stephanie S and Cvach, Maria and Dawson, Patricia B and Straus, Harriet and Hill, Elizabeth E},
  journal={Journal of nursing care quality},
  volume={22},
  number={4},
  pages={293--298},
  year={2007},
  publisher={LWW}
}

@article{rajkomar2019machine,
  title={Machine learning in medicine},
  author={Rajkomar, Alvin and Dean, Jeffrey and Kohane, Isaac},
  journal={New England Journal of Medicine},
  volume={380},
  number={14},
  pages={1347--1358},
  year={2019},
  publisher={Mass Medical Soc}
}

@inproceedings{rennie2005loss,
  title={Loss functions for preference levels: Regression with discrete ordered labels},
  author={Rennie, Jason DM and Srebro, Nathan},
  booktitle={Proceedings of the IJCAI multidisciplinary workshop on advances in preference handling},
  volume={1},
  pages={1--6},
  year={2005},
  organization={AAAI Press, Menlo Park, CA}
}

@article{rennie2005smooth,
  title={Smooth hinge classification},
  author={Rennie, Jason DM},
  journal={Proceeding of Massachusetts Institute of Technology},
  year={2005}
}

@article{shashua2002ranking,
  title={Ranking with large margin principle: Two approaches},
  author={Shashua, Amnon and Levin, Anat},
  journal={Advances in neural information processing systems},
  volume={15},
  year={2002}
}

@article{ustun2019learning,
  title={Learning optimized risk scores},
  author={Ustun, Berk and Rudin, Cynthia},
  journal={Journal of Machine Learning Research},
  volume={20},
  number={150},
  pages={1--75},
  year={2019}
}

@article{holzinger2017we,
  title={What do we need to build explainable AI systems for the medical domain?},
  author={Holzinger, Andreas and Biemann, Chris and Pattichis, Constantinos S and Kell, Douglas B},
  journal={arXiv preprint arXiv:1712.09923},
  year={2017}
}

@inproceedings{wang2019adaptive,
  title={Adaptive graph guided disambiguation for partial label learning},
  author={Wang, Deng-Bao and Li, Li and Zhang, Min-Ling},
  booktitle={Proceedings of the 25th ACM SIGKDD international conference on knowledge discovery \& data mining},
  pages={83--91},
  year={2019}
}

@inproceedings{zhang2015solving,
  title={Solving the partial label learning problem: An instance-based approach.},
  author={Zhang, Min-Ling and Yu, Fei},
  booktitle={IJCAI},
  pages={4048--4054},
  year={2015}
}

@article{karapanagiotis2023tailored,
  title={Tailored Bayes: a risk modeling framework under unequal misclassification costs},
  author={Karapanagiotis, Solon and Benedetto, Umberto and Mukherjee, Sach and Kirk, Paul DW and Newcombe, Paul J},
  journal={Biostatistics},
  volume={24},
  number={1},
  pages={85--107},
  year={2023},
  publisher={Oxford University Press}
}

@article{peterson1990partial,
  title={Partial proportional odds models for ordinal response variables},
  author={Peterson, Bercedis and Harrell Jr, Frank E},
  journal={Journal of the Royal Statistical Society: Series C (Applied Statistics)},
  volume={39},
  number={2},
  pages={205--217},
  year={1990},
  publisher={Wiley Online Library}
}

@article{uney_mixed-integer_2006,
	title = {A mixed-integer programming approach to multi-class data classification problem},
	volume = {173},
	issn = {0377-2217},
	url = {https://www.sciencedirect.com/science/article/pii/S0377221705006909},
	doi = {10.1016/j.ejor.2005.04.049},
	pages = {910--920},
	number = {3},
    year = {2006},
	journal = {European Journal of Operational Research},
	shortjournal = {European Journal of Operational Research},
	author = {Üney, Fadime and Türkay, Metin},
	urldate = {2025-08-07},
	date = {2006-09-16},
}

@article{bertsimas_classification_2007,
	title = {Classification and Regression via Integer Optimization},
	volume = {55},
	issn = {0030-364X},
	url = {https://pubsonline.informs.org/doi/abs/10.1287/opre.1060.0360},
	doi = {10.1287/opre.1060.0360},
	pages = {252--271},
	number = {2},
    year = {2007},
	journal = {Operations Research},
	author = {Bertsimas, Dimitris and Shioda, Romy},
	urldate = {2025-08-07},
	date = {2007-04},
	note = {Publisher: {INFORMS}},
}

@article{dedieu_learning_2021,
	title = {Learning sparse classifiers: continuous and mixed integer optimization perspectives},
	volume = {22},
	issn = {1532-4435},
	shorttitle = {Learning sparse classifiers},
	pages = {135:6008--135:6054},
	number = {1},
    year = {2021}, 
	journal = {Journal of Machine Learning Ressearch},
	author = {Dedieu, Antoine and Hazimeh, Hussein and Mazumder, Rahul},
	date = {2021-01-01},
}

@article{billiet_interval_2018,
	title = {Interval Coded Scoring: a toolbox for interpretable scoring systems},
	volume = {4},
	rights = {http://creativecommons.org/licenses/by/4.0/},
	issn = {2376-5992},
	url = {https://peerj.com/articles/cs-150},
	doi = {10.7717/peerj-cs.150},
	shorttitle = {Interval Coded Scoring},
	pages = {e150},
    year = {2018}, 
	journal = {{PeerJ} Computer Science},
	author = {Billiet, Lieven and Van Huffel, Sabine and Van Belle, Vanya},
	urldate = {2025-11-07},
	date = {2018-04-02},
	langid = {english},
}

@article{moreno_sequential_2023,
	title = {The Sequential Organ Failure Assessment ({SOFA}) Score: has the time come for an update?},
	volume = {27},
	issn = {1466-609X},
	doi = {10.1186/s13054-022-04290-9},
	shorttitle = {The Sequential Organ Failure Assessment ({SOFA}) Score},
	pages = {15},
	number = {1},
    year = {2023},
	journal = {Critical Care (London, England)},
	shortjournal = {Crit Care},
	author = {Moreno, Rui and Rhodes, Andrew and Piquilloud, Lise and Hernandez, Glenn and Takala, Jukka and Gershengorn, Hayley B. and Tavares, Miguel and Coopersmith, Craig M. and Myatra, Sheila N. and Singer, Mervyn and Rezende, Ederlon and Prescott, Hallie C. and Soares, Márcio and Timsit, Jean-François and de Lange, Dylan W. and Jung, Christian and De Waele, Jan J. and Martin, Greg S. and Summers, Charlotte and Azoulay, Elie and Fujii, Tomoko and {McLean}, Anthony S. and Vincent, Jean-Louis},
	date = {2023-01-13},
	pmid = {36639780},
	pmcid = {PMC9837980},
}

@article{vincent_sofa_1996,
	title = {The {SOFA} (Sepsis-related Organ Failure Assessment) score to describe organ dysfunction/failure. On behalf of the Working Group on Sepsis-Related Problems of the European Society of Intensive Care Medicine},
	volume = {22},
	issn = {0342-4642},
	doi = {10.1007/BF01709751},
	pages = {707--710},
	number = {7},
    year = {1996},
	journal = {Intensive Care Medicine},
	shortjournal = {Intensive Care Med},
	author = {Vincent, J. L. and Moreno, R. and Takala, J. and Willatts, S. and De Mendonça, A. and Bruining, H. and Reinhart, C. K. and Suter, P. M. and Thijs, L. G.},
	date = {1996-07},
	pmid = {8844239},
}

\end{document}